%% file: main.tex
\documentclass[10pt,twocolumn,letterpaper]{article}

\usepackage{iccv}
\usepackage{times}
\usepackage{epsfig}
\usepackage{graphicx}
\usepackage{amsmath}
\usepackage{amssymb}
\usepackage[numbers, sort]{natbib}

\usepackage[pagebackref=true,breaklinks=true,colorlinks,bookmarks=false]{hyperref}

\usepackage{helvet}   
\usepackage{courier}  
\usepackage{url}      
\usepackage{xcolor}
\usepackage{booktabs}
\usepackage{multirow}
\usepackage{subcaption}

\usepackage[export]{adjustbox}

\DeclareMathAlphabet      {\mathbfit}{OML}{cmm}{b}{it}

\newcommand{\ra}[1]{\renewcommand{\arraystretch}{#1}}

\usepackage{pifont}

\renewcommand{\etal}{\textit{et~al}\mbox{.}}
\renewcommand{\eg}{e.g.,\ }
\renewcommand{\ie}{i.e.,\ }
\newcommand{\bbR}{{\mathbb{R}}}

\newlength{\threeimg}

\newlength\paramargin
\newlength\figmargin
\newlength\tablemargin
\newlength\secmargin
\newlength\figcapmargin
\newlength\tablecapmargin

\iccvfinalcopy 

\def\iccvPaperID{157} 

\ificcvfinal\fi
\begin{document}

\title{Recover and Identify:\\
A Generative Dual Model for Cross-Resolution Person Re-Identification}

\author{
%
Yu-Jhe~Li$^{1,3}$\thanks{~indicates equal contributions.}
\hspace{4.0mm}
Yun-Chun~Chen$^{1,2,3*}$
\hspace{4.0mm}
Yen-Yu~Lin$^{2}$
\hspace{4.0mm}
Xiaofei~Du$^{5}$
\hspace{4.0mm}
Yu-Chiang~Frank~Wang$^{1,3,4}$
\\
%
%
$^{1}$National Taiwan University
\hspace{6.0mm}
$^{2}$Academia Sinica
\\
$^{3}$MOST Joint Research Center for AI Technology and All Vista Healthcare
\\
$^{4}$Asus Intelligent Cloud Services
\hspace{6.0mm}
$^{5}$Umbo Computer Vision
\\
{\tt\small \{\url{yujheli}, \url{ycchen}, \url{ycwang}\}\url{@ntu.edu.tw}, \url{yylin@citi.sinica.edu.tw}, \url{xiaofei.du@umbocv.com}}
}

\maketitle
\thispagestyle{empty}

\input{abstract}

\input{introduction}

\input{related-work}

\input{method}

\input{experiment}

\input{conclusion}

\vspace{\paramargin}
\paragraph{Acknowledgements.} This work is supported in part by Ministry of Science and Technology (MOST) under grants 107-2628-E-001-005-MY3, 108-2634-F-007-009, and 108-2634-F-002-018, and Umbo Computer Vision.

{\small
\bibliographystyle{ieee_fullname}
\bibliography{egbib}
}

\end{document}

%% file: abstract.tex
\begin{abstract}

Person re-identification (re-ID) aims at matching images of the same identity across camera views. Due to varying distances between cameras and persons of interest, resolution mismatch can be expected, which would degrade person re-ID performance in real-world scenarios. To overcome this problem, we propose a novel generative adversarial network to address cross-resolution person re-ID, allowing query images with varying resolutions. By advancing adversarial learning techniques, our proposed model learns resolution-invariant image representations while being able to recover the missing details in low-resolution input images. The resulting features can be jointly applied for improving person re-ID performance due to preserving resolution invariance and recovering re-ID oriented discriminative details. Our experiments on five benchmark datasets confirm the effectiveness of our approach and its superiority over the state-of-the-art methods, especially when the input resolutions are unseen during training.

\end{abstract}

%% file: introduction.tex
\section{Introduction}

Person re-identification (re-ID)~\cite{zheng2016person} aims at recognizing the same person across images taken by different cameras, and is an active research topic in computer vision. A variety of applications ranging from person tracking~\cite{andriluka2008people}, video surveillance system~\cite{khan2016person}, to computational forensics~\cite{vezzani2013people} are highly correlated this research topic. Nevertheless, due to the presence of background clutter, occlusion, illumination or viewpoint changes, person re-ID remains a challenging task for practical applications.

\begin{figure}[t]
  \centering\includegraphics[width=\linewidth]{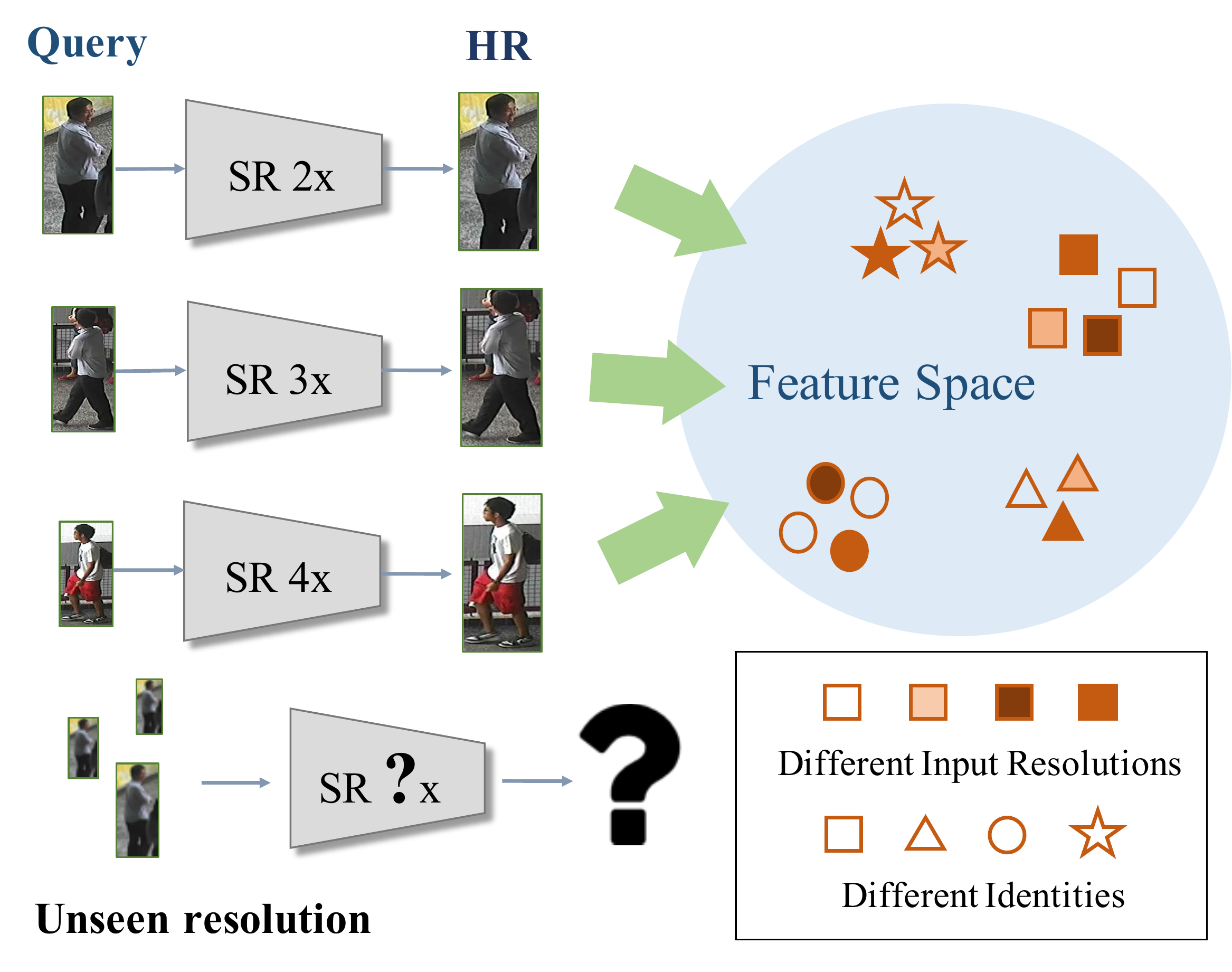}
  \vspace{-6.0mm}
  \caption{\textbf{Illustration and challenges of cross-resolution person re-ID.} Note that existing approaches typically leverage SR models with pre-selected resolutions followed by person re-ID modules. This cannot not be easily applied to query images with varying or unseen resolutions.}
  \vspace{-4.0mm}
  \label{fig:teaser}
\end{figure}

Driven by the recent success of convolutional neural networks (CNNs), several learning-based methods~\cite{lin2017improving,hermans2017defense,zhong2017camera,si2018dual} have been proposed. Despite promising performances, these methods are typically developed under the assumption that both query and gallery images are of \emph{similar} or \emph{sufficiently high} resolutions. This assumption, however, may not hold in practice since image resolutions would vary drastically. For instance, query images captured by surveillance cameras are often of low resolution (LR) whereas those in the gallery set are carefully selected beforehand and are of high resolution (HR). As a result, direct matching of LR query images and HR gallery ones would lead to non-trivial \emph{resolution mismatch} problems.

To address cross-resolution person re-ID, most existing methods~\cite{wang2018cascaded,jiao2018deep} employ super-resolution (SR) models to convert LR inputs into their HR versions followed by person re-ID. However, these methods suffer from two limitations. First, each employed SR model is designed to upscale image resolutions by a particular factor. Thus, these methods need to \emph{pre-determine} the resolutions of LR queries so that the corresponding SR models can be applied. However, designing SR models for each possible resolution input makes these methods hard to scale. Second, in the real-world scenario, queries can be with \emph{various} resolutions even with the resolutions that are \emph{unseen} during training. As illustrated in Figure~\ref{fig:teaser}, queries with varying or unseen resolutions would restrict the applicability of the person re-ID methods that employ SR models since one cannot assume the resolutions of the input images will be known in advance.

In this paper, we propose \emph{Cross-resolution Adversarial Dual Network} (CAD-Net) for cross-resolution person re-ID. The key characteristics of CAD-Net are two-fold. First, to address the resolution variations, CAD-Net derives the \emph{resolution-invariant representations} via adversarial learning. This allows our model to handle images of \emph{varying} and even \emph{unseen} resolutions. Second, CAD-Net learns to recover the missing details in LR input images. Together with the resolution-invariant features, our model generates HR images \emph{preferable for person re-ID}, achieving the state-of-the-art performance on cross-resolution person re-ID. It is worth noting that the above image resolution recovery and cross-resolution person re-ID are realized by a \emph{single} model learned in an \emph{end-to-end} fashion. 
%

The contributions of this paper are highlighted below: 

\begin{itemize}
  \vspace{-1.5mm} 
  \item We propose an end-to-end trainable network which advances adversarial learning strategies for cross-resolution person re-ID.

  \vspace{-1.5mm} 
  \item Our model learns resolution-invariant representations while recovering the missing details in LR input images, resulting in improved cross-resolution person re-ID performance.

  \vspace{-1.5mm} 
  \item Our model is able to handle query images with varying or even unseen resolutions without the need to pre-determine the input resolutions.
  
  \vspace{-1.5mm} 
  \item Extensive experimental results on five challenging datasets confirm that our method performs favorably against the state-of-the-art person re-ID approaches.
\end{itemize}

%% file: related-work.tex
\section{Related Work}

\paragraph{Person re-ID.}

A variety of existing methods~\cite{lin2017improving,shen2018deep,shen2018person,kalayeh2018human,cheng2016person,chang2018multi,chen2018group} are developed to address various challenges in person re-ID, such as background clutter, viewpoint changes, and pose variations. For instance, Yang~\etal~\cite{zhong2017camera} learn a camera-invariant subspace to deal with the style variations caused by different cameras. Liu~\etal~\cite{liu2018pose} develop a pose-transferable framework based on the generative adversarial network (GAN)~\cite{goodfellow2014generative} to yield pose-specific images for tackling the pose variations. Several methods~\cite{li2018harmonious,song2018mask,si2018dual} addressing background clutter leverage attention mechanisms~\cite{chen2018deep,chen2019saliency,chen2019show,lin2018learning} to emphasize the discriminative parts. Another research trend focuses on domain adaptation~\cite{hoffman2017cycada,chen2019crdoco} for person re-ID~\cite{wei2018person,image-image18}. By viewing image-to-image translation methods as a data augmentation technique, these methods employ image translation modules, \eg CycleGAN~\cite{zhu2017unpaired}, to generate viewpoint specific images with labels. However, the above approaches typically assume that both query and gallery images are of similar or sufficiently high resolutions, which might not be practical for real-world applications.

\vspace{\paramargin}
\paragraph{Cross-resolution person re-ID.}

A number of methods~\cite{li2015multi,jing2015super,wang2016scale,jiao2018deep,wang2018cascaded,RAIN} have been proposed to address the problem of resolution mismatch in person re-ID. Li~\etal~\cite{li2015multi} jointly perform multi-scale distance metric learning and cross-scale image domain alignment. Jing~\etal~\cite{jing2015super} develop a semi-coupled low-rank dictionary learning framework to seek a mapping between HR and LR images. Wang~\etal~\cite{wang2016scale} learn a discriminating scale-distance function space by varying the image scale of LR images when matching with the HR ones. Nevertheless, these methods adopt hand-crafted descriptors, which cannot easily adapt the developed models to the tasks of interest, and thus may lead to sub-optimal person re-ID performance.

Recently, three CNN-based methods~\cite{jiao2018deep,wang2018cascaded,RAIN} are presented for cross-resolution person re-ID. The network of SING~\cite{jiao2018deep} is composed of several SR sub-networks and a person re-ID module to carry out LR person re-ID. On the other hand, CSR-GAN~\cite{wang2018cascaded} cascades multiple SR-GANs~\cite{ledig2017photo} and progressively recovers the details of LR images to address the resolution mismatch problem. In spite of their promising results, such methods require the training of pre-defined SR models. As mentioned earlier, the degree of resolution mismatch, \ie the resolution difference between the query and gallery images, is typically \emph{unknown beforehand}. Moreover, if the resolution of the input LR query is unseen during training, the above methods cannot be easily applied or might not lead to satisfactory performance. Apart from these methods, RAIN~\cite{RAIN} aligns the feature distributions of HR and LR images, showing some performance improvement over existing algorithms.

Similar to RAIN~\cite{RAIN}, our method also performs feature distribution alignment between HR and LR images. Our model differs from RAIN~\cite{RAIN} in two aspects. First, our model derives resolution-invariant representations and recovers the missing details in LR input images. By jointly considering features of both modalities, our algorithm further improves the performance. Second, the HR image recovery is learned in an end-to-end fashion, allowing our model to recover HR images preferable for person re-ID. Experimental results demonstrate that our approach can be applied to input images of varying and even unseen resolutions using only a single model.

\begin{figure*}[t]
  \centering
  \includegraphics[width=\linewidth]{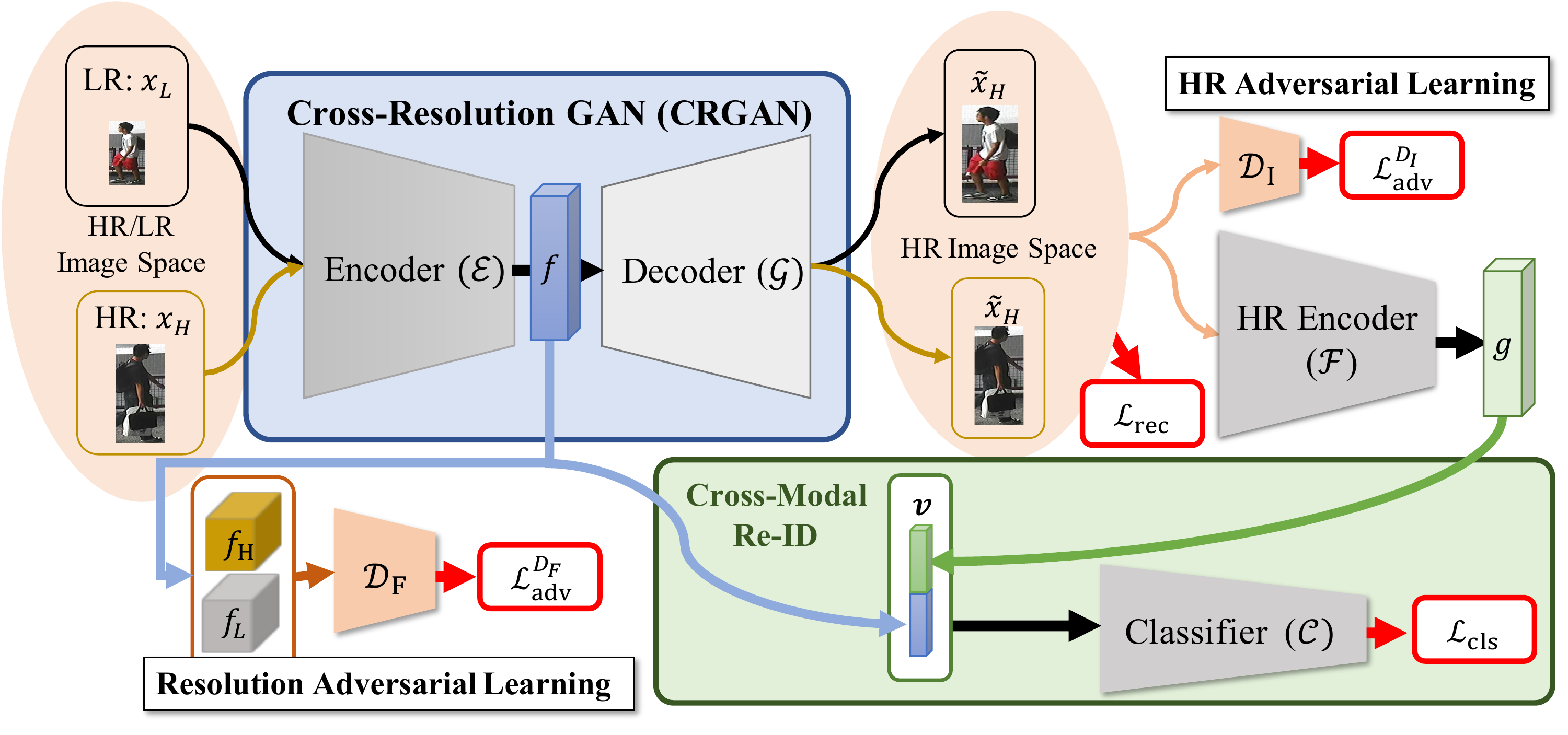}
  \vspace{-7.5mm}
  \caption{\textbf{Overview of Cross-resolution Adversarial Dual Network (CAD-Net).} CAD-Net comprises Cross-Resolution GAN (CRGAN) and Cross-Modal Re-ID network. The former learns resolution-invariant representations and recovers the missing details in LR input images, while the latter considers both feature modalities for cross-resolution person re-ID.}
  \vspace{-3.0mm}
  \label{fig:Model}
\end{figure*}

\vspace{\paramargin}
\paragraph{Cross-resolution vision applications.}

The issues regarding cross-resolution handling have been studied in the literature. For face recognition, existing approaches typically rely on face hallucination algorithms~\cite{zhu2016deep,yu2017hallucinating} or SR mechanisms~\cite{kim2016accurate,dahl2017pixel,dong2016image} to super-resolve the facial details. Unlike the above existing methods that focus on synthesizing the facial details, our model learns to recover re-ID oriented discriminative details. Together with the derived resolution-invariant features, our model would considerably boost the person re-ID performance while allowing query images with varying and even unseen resolutions.

%% file: method.tex
\section{Proposed Method}

In this section, we first provide an overview of our proposed approach. We then describe the details of each network component as well as the loss functions.

\subsection{Algorithmic Overview}

We define the notations to be used in this paper. In the training stage, we have access to a set of~$N$ HR images $X_H = \{x_i^H\}_{i=1}^N$ and its corresponding label set $Y_H = \{y_i^H\}_{i=1}^N$, where $x_i^H \in \bbR^{H \times W \times 3}$ and $y_i^H \in \bbR$ are the $i^\mathrm{th}$ HR image and its label, respectively. To allow our model to handle images of different resolutions, we generate a \emph{synthetic} LR image set $X_L = \{x_i^L\}_{i=1}^N$ by down-sampling each image in $X_H$, followed by resizing them back to the original image size via bilinear up-sampling (\ie $x_i^L \in \bbR^{H \times W \times 3}$), where $x_i^L$ is the synthetic LR image of $x_i^H$. Obviously, the label set $Y_L$ for $X_L$ is identical to $Y_H$.

As shown in Figure~\ref{fig:Model}, our network comprises two components: Cross-Resolution Generative Adversarial Network (CRGAN) and Cross-Modal Re-ID network. To achieve cross-resolution person re-ID, our CRGAN simultaneously learns a resolution-invariant representation $f \in \bbR^{h \times w \times d}$ ($h \times w$ is the spatial size of $f$ whereas $d$ denotes the number of channels) from the input cross-resolution images, while producing the associated HR images as the decoder outputs. The recovered HR output image will be encoded as an HR representation $g \in \bbR^{h \times w \times d}$ by the HR encoder. For person re-ID, we first concatenate $f$ and $g$ to form a joint representation $\mathbfit{v} = [f, g] \in \bbR^{h \times w \times 2d}$. The classifier then takes the joint representation $\mathbfit{v}$ as input to perform person identity classification. The details of each component are elaborated in the following subsections.

As for testing, our network takes a query image resized to $H \times W \times 3$ as the input, and computes the joint representation $\mathbfit{v} = [f, g] \in \bbR^{h \times w \times 2d}$. We then apply global average pooling ($\mathrm{GAP}$) to $\mathbfit{v}$ for deriving a joint feature vector $\mathbfit{u} = \mathrm{GAP}(\mathbfit{v}) \in \bbR^{2d}$, which is applied to match the gallery images via nearest neighbor search with Euclidean distance. It is worth repeating that, the query image during testing can be with varying resolutions or with unseen ones during training (verified in experiments).

\subsection{Cross-Resolution GAN (CRGAN)}

In CRGAN, we have a cross-resolution encoder $\mathcal{E}$ which converts input images across different resolutions into resolution-invariant representations, followed by a high-resolution decoder $\mathcal{G}$ recovering the associated HR versions.

\vspace{\paramargin}
\paragraph{Cross-resolution encoder $\mathcal{E}$.}

Since our goal is to perform cross-resolution person re-ID, we encourage the cross-resolution encoder $\mathcal{E}$ to extract resolution-invariant features for input images across resolutions (\eg HR images $X_H$ and LR ones $X_L$). To achieve this, we advance adversarial learning strategies and deploy a resolution discriminator $\mathcal{D}_{F}$ in the latent \emph{feature space}. This discriminator $\mathcal{D}_{F}$ takes the feature maps $f_H$ and $f_L$ as inputs to determine whether the input feature maps are from $X_H$ or $X_L$. To be more precise, we define the feature-level adversarial loss $\mathcal{L}_\mathrm{adv}^{\mathcal{D}_{F}}$ as
\begin{equation}
  \begin{aligned}
  \mathcal{L}_\mathrm{adv}^{\mathcal{D}_{F}} = &~ \mathbb{E}_{x_H \sim X_H}[\log(\mathcal{D}_{F}(f_H))]\\
  + &~ \mathbb{E}_{x_L \sim X_L}[\log(1 - \mathcal{D}_{F}(f_L))],
  \end{aligned}
  \label{eq:adv_loss_feature}
\end{equation}
where $f_H = \mathcal{E}({x_H})$ and $f_L = \mathcal{E}({x_L}) \in \bbR^{h \times w \times d}$ denote the encoded HR and LR image features, respectively.\footnote{For simplicity, we omit the subscript $i$, denote HR and LR images as $x_H$ and $x_L$, and represent their corresponding labels as $y_H$ and $y_L$.}

With loss $\mathcal{L}_\mathrm{adv}^{\mathcal{D}_{F}}$, our resolution discriminator $\mathcal{D}_{F}$ aligns the feature distributions across resolutions, carrying out the learning of resolution-invariant representations.

\vspace{\paramargin}
\paragraph{High-resolution decoder $\mathcal{G}$.}

In addition to learning the resolution-invariant representation $f$, our CRGAN further synthesizes the associated HR images. This is to recover the missing details in LR input images, together with the person re-ID task to be performed later in the cross-modal re-ID network.

To achieve this goal, we have an HR decoder $\mathcal{G}$ in our CRGAN which reconstructs (or recovers) the HR images as the outputs. To accomplish this, we apply an HR reconstruction loss $\mathcal{L}_\mathrm{rec}$ between the reconstructed HR images and their corresponding HR ground-truth images. Specifically, the HR reconstruction loss $\mathcal{L}_\mathrm{rec}$ is defined as
\begin{equation}
  \label{eq:rec}
  \begin{aligned}
  \mathcal{L}_\mathrm{rec} = &~ \mathbb{E}_{x_H \sim X_H}[\|\mathcal{G}(f_H) - x_H\|_1]\\
  + &~ \mathbb{E}_{x_L \sim X_L}[\|\mathcal{G}(f_L) - x_{H}\|_1],
  \end{aligned}
\end{equation}
where the HR ground-truth image associated with $x_L$ is $x_H$. Following Huang~\etal~\cite{huang2018munit}, we adopt the $\ell_1$ norm in the loss $\mathcal{L}_\mathrm{rec}$ as it preserves image sharpness. We note that both $X_H$ and $X_L$ will be shuffled during training. That is, images of the same identity but different resolutions will not necessarily be observed by CRGAN at the same time.

It is worth noting that, while the aforementioned HR reconstruction loss $\mathcal{L}_\mathrm{rec}$ could reduce information loss in the latent feature space, we follow Ledig~\etal~\cite{ledig2017photo} and introduce skip connections between the cross-resolution encoder $\mathcal{E}$ and the HR decoder $\mathcal{G}$. This would facilitate the learning process of image reconstruction, as well as allowing more efficient gradient propagation.

To encourage the HR decoder $\mathcal{G}$ to produce more perceptually realistic HR outputs and associate with the task of person re-ID, we further adopt adversarial learning in the \emph{image space} and introduce an HR image discriminator $\mathcal{D}_{I}$ which takes the recovered HR images (\ie $\mathcal{G}(f_L)$ and $\mathcal{G}(f_H)$) and their corresponding HR ground-truth images as inputs to distinguish whether the input images are real or fake~\cite{ledig2017photo,wang2018cascaded}. Specifically, we define the image-level adversarial loss $\mathcal{L}_\mathrm{adv}^{\mathcal{D}_{I}}$ as
\begin{equation}\scriptsize
  \begin{aligned}
  \mathcal{L}_\mathrm{adv}^{\mathcal{D}_{I}} = &~ \mathbb{E}_{x_H \sim X_H}[\log(\mathcal{D}_{I}(x_H))] + \mathbb{E}_{x_L \sim X_L}[\log(1 - \mathcal{D}_{I}(\mathcal{G}(f_L)))] \\
  + &~ \mathbb{E}_{x_H \sim X_H}[\log(\mathcal{D}_{I}(x_H))] + \mathbb{E}_{x_H \sim X_H}[\log(1 - \mathcal{D}_{I}(\mathcal{G}(f_H)))].
  \end{aligned}
  \label{eq:adv_loss_image}
\end{equation}

It is also worth repeating that the goal of this HR decoder $\mathcal{G}$ is not simply to recover the missing details in LR input images, but also to have such recovered HR images aligned with the learning task of interest (\ie person re-ID). Namely, we encourage the HR decoder $\mathcal{G}$ to perform \textit{re-ID oriented} HR recovery, which is further realized by the following cross-modal re-ID network.

\subsection{Cross-Modal Re-ID}

As shown in Figure~\ref{fig:Model}, the cross-modal re-ID network first applies an HR encoder $\mathcal{F}$, which takes the reconstructed HR image from CRGAN as input, to derive the HR feature representation $g \in \bbR^{h \times w \times d}$. Then, a classifier $\mathcal{C}$ is learned to complete person re-ID.

As for the input to the classifier $\mathcal{C}$, we jointly consider the feature representations of two different modalities for person identity classification, \ie the resolution-invariant representation $f$ and the HR representation $g$. The former preserves content information, while the latter observes the recovered HR details for person re-ID. Thus, we have the classifier $\mathcal{C}$ take the concatenated feature representation $\mathbfit{v} = [f, g] \in \bbR^{h \times w \times 2d}$ as the input. In this work, the adopted classification loss $\mathcal{L}_\mathrm{cls}$ is the integration of the identity loss $\mathcal{L}_\mathrm{id}$ and the triplet loss $\mathcal{L}_\mathrm{tri}$~\cite{hermans2017defense}, and is defined as
\begin{equation}
  \begin{aligned}
  \mathcal{L}_\mathrm{cls} = \mathcal{L}_\mathrm{id} + \mathcal{L}_\mathrm{tri},
  \end{aligned}
  \label{eq:cls}
\end{equation}
where the identity loss $\mathcal{L}_\mathrm{id}$ computes the softmax cross entropy between the classification prediction and the corresponding ground-truth one hot vector, while the triplet loss $\mathcal{L}_\mathrm{tri}$ is introduced to enhance the discrimination ability during person re-ID process and is defined as
\begin{equation}
  \begin{aligned}
  \mathcal{L}_\mathrm{tri}
   = &~ \mathbb{E}_{(x_H,y_H) \sim (X_H,Y_H)}\max(0, \phi + d_\mathrm{pos}^H - d_\mathrm{neg}^H) \\
  + &~ \mathbb{E}_{(x_L,y_L) \sim (X_L,Y_L)}\max(0, \phi + d_\mathrm{pos}^L - d_\mathrm{neg}^L),
  \end{aligned}
  \label{eq:tri}
\end{equation}
where $d_\mathrm{pos}$ and $d_\mathrm{neg}$ are the distances between the positive (same label) and the negative (different labels) image pairs, respectively, and $\phi > 0$ serves as the margin. We note that weighted identity classification loss~\cite{chen2017deep} can also be adopted to improve person identity classification.

It can be seen that the above cross-resolution person re-ID framework is very different from existing one like CSR-GAN~\cite{wang2018cascaded}, which addresses SR and person re-ID \emph{separately}. More importantly, the aforementioned identity loss $\mathcal{L}_\mathrm{id}$ not only updates the classifier $\mathcal{C}$, but also refines the HR decoder $\mathcal{G}$ in our CRGAN. This is the reason why our CRGAN is able to produce \emph{re-ID oriented} HR outputs, \ie the recovered HR details preferable for person re-ID.

\vspace{\paramargin}
\paragraph{Full objective.}

The total loss function $\mathcal{L}$ for training our proposed CAD-Net is summarized as follows:
\begin{equation}
  \begin{split}
  \mathcal{L} = \mathcal{L}_\mathrm{cls} + \lambda_\mathrm{adv}^{\mathcal{D}_{F}}\cdot\mathcal{L}_\mathrm{adv}^{\mathcal{D}_{F}} + \lambda_\mathrm{rec}\cdot\mathcal{L}_\mathrm{rec} + \lambda_\mathrm{adv}^{\mathcal{D}_{I}}\cdot\mathcal{L}_\mathrm{adv}^{\mathcal{D}_{I}},
  \end{split}
  \label{eq:fullobj}
\end{equation}
where $\lambda_\mathrm{adv}^{\mathcal{D}_{F}}$, $\lambda_\mathrm{rec}$, and $\lambda_\mathrm{adv}^{\mathcal{D}_{I}}$ are the hyper-parameters used to control the relative importance of the corresponding losses. We note that losses $\mathcal{L}_\mathrm{adv}^{\mathcal{D}_{F}}$, $\mathcal{L}_\mathrm{rec}$, and $\mathcal{L}_\mathrm{adv}^{\mathcal{D}_{I}}$ are developed to learn CRGAN, while loss $\mathcal{L}_\mathrm{cls}$ is designed to update both CRGAN and cross-modal re-ID network.

To train our network using training HR images and their down-sampled LR ones, we minimize the HR reconstruction loss $\mathcal{L}_\mathrm{rec}$ for updating our CRGAN, and the classification loss $\mathcal{L}_\mathrm{cls}$ for jointly updating CRGAN and cross-modal re-ID network. The image-level adversarial loss $\mathcal{L}_{adv}^{\mathcal{D}_I}$ is computed for producing perceptually realistic HR images while the feature-level adversarial loss $\mathcal{L}_{adv}^{\mathcal{D}_F}$ is optimized for learning resolution-invariant representations.

%% file: experiment.tex
\section{Experiments}

We first provide the implementation details, followed by dataset descriptions and settings. Both quantitative and qualitative results are presented, including ablation studies.

\subsection{Implementation Details}

We implement our model using PyTorch. ResNet-$50$~\cite{he2016deep} pretrained on ImageNet is used to build the cross-resolution encoder $\mathcal{E}$ and the HR encoder $\mathcal{F}$. Note that since $\mathcal{E}$ and $\mathcal{F}$ work for different tasks, these two components do not share weights. The classifier $\mathcal{C}$ is composed of a global average pooling layer and a fully connected layer followed a softmax activation. The architecture of the resolution discriminator $\mathcal{D}_F$ is the same as that adopted by Tsai~\etal~\cite{tsai2018learning}. The structure of the HR image discriminator $\mathcal{D}_I$ is similar to ResNet-$18$~\cite{he2016deep}. Our HR decoder $\mathcal{G}$ is similar to that proposed by Miyato~\etal~\cite{miyato2018cgans}. Components $\mathcal{D}_F$, $\mathcal{D}_I$, $\mathcal{G}$, and $\mathcal{C}$ are all randomly initialized. We use stochastic gradient descent to train the proposed model. For components $\mathcal{E}$, $\mathcal{G}$, $\mathcal{F}$, and $\mathcal{C}$, the learning rate, momentum, and weight decay are $1 \times 10^{-3}$, $0.9$, and $5 \times 10^{-4}$, respectively. For the two discriminators $\mathcal{D}_F$ and $\mathcal{D}_I$, the learning rate is set to $1 \times 10^{-4}$. The batch size is $32$. The margin $\phi$ in the triplet loss $\mathcal{L}_\mathrm{tri}$ is set to $2$. We set the hyper-parameters in all the experiments as follows: $\lambda_\mathrm{adv}^{\mathcal{D}_{F}} = 1$, $\lambda_\mathrm{rec} = 1$, and $\lambda_\mathrm{adv}^{\mathcal{D}_{I}} = 1$. All images of various resolutions are resized to $256 \times 128 \times 3$ in advance. We train our model on a single NVIDIA GeForce GTX $1080$ GPU with $12$ GB memory.


\subsection{Datasets}

We evaluate the proposed method on five datasets, each of which is described as follows.

\vspace{\paramargin}
\paragraph{CUHK03~\cite{li2014deepreid}.} The CUHK03 dataset comprises $14,097$ images of $1,467$ identities with $5$ different camera views. Following CSR-GAN~\cite{wang2018cascaded}, we use the $1,367/100$ training/test identity split.

\vspace{\paramargin}
\paragraph{VIPeR~\cite{gray2008viewpoint}.} The VIPeR dataset contains $632$ person-image pairs captured by $2$ cameras. Following SING~\cite{jiao2018deep}, we randomly divide this dataset into two non-overlapping halves based on the identity labels. Namely, images of a subject belong to either the training set or the test set.

\vspace{\paramargin}
\paragraph{CAVIAR~\cite{Cheng:BMVC11}.} The CAVIAR dataset is composed of $1,220$ images of $72$ person identities captured by $2$ cameras. Following SING~\cite{jiao2018deep}, we discard $22$ people who only appear in the closer camera, and split this dataset into two non-overlapping halves according to the identity labels.

\vspace{\paramargin}
\paragraph{Market-$1501$~\cite{zheng2015scalable}.} The Market-$1501$ dataset consists of $32,668$ images of $1,501$ identities with $6$ camera views. We use the widely adopted $751/750$ training/test identity split.

\vspace{\paramargin}
\paragraph{DukeMTMC-reID~\cite{zheng2017unlabeled}.} The DukeMTMC-reID dataset contains $36,411$ images of $1,404$ identities captured by $8$ cameras. We adopt the benchmarking $702/702$ training/test identity split.

\subsection{Experimental Settings and Evaluation Metrics}



We evaluate the proposed method using {\em cross-resolution person re-ID} setting~\cite{jiao2018deep} where the test (query) set is composed of LR images while the gallery set contains HR images only. In all of the experiments, we adopt the standard single-shot person re-ID setting~\cite{jiao2018deep,liao2015person} and use the average cumulative match characteristic as the evaluation metric.

\input{exp/exp-ReID.tex}

\subsection{Evaluation and Comparisons}

Following SING~\cite{jiao2018deep}, we consider multiple low-resolution (MLR) person re-ID and evaluate the proposed method on \emph{four synthetic} and \emph{one real-world} benchmarks. To construct the synthetic MLR datasets (\ie MLR-CUHK03, MLR-VIPeR, MLR-Market-$1501$, and MLR-DukeMTMC-reID), we follow SING~\cite{jiao2018deep} and down-sample images taken by one camera by a randomly selected down-sampling rate $r \in \{2, 3, 4\}$ (\ie the size of the down-sampled image becomes $\frac{H}{r} \times \frac{W}{r} \times 3$), while the images taken by the other camera(s) remain unchanged. The CAVIAR dataset inherently contains realistic images of multiple resolutions, and is a \emph{genuine} and more challenging dataset for evaluating MLR person re-ID.

We compare our approach with methods developed for cross-resolution person re-ID, including JUDEA~\cite{li2015multi}, SLD$^2$L~\cite{jing2015super}, SDF~\cite{wang2016scale}, SING~\cite{jiao2018deep}, and CSR-GAN~\cite{wang2018cascaded}, and methods developed for standard person re-ID, including CamStyle~\cite{zhong2017camera} and FD-GAN~\cite{ge2018fd}. For methods developed for cross-resolution person re-ID, the training set contains HR images and LR ones with all three down-sampling rates $r \in \{2, 3, 4\}$ for each person. For methods developed for standard person re-ID, the training set contains HR images for each identity only.

Table~\ref{table:exp-ReID} reports the quantitative results recorded at ranks $1$, $5$, and $10$ on all five adopted datasets. For CSR-GAN~\cite{wang2018cascaded} on MLR-CUHK03, CAVIAR, MLR-Market-$1501$, and MLR-DukeMTMC-reID, and CamStyle~\cite{zhong2017camera} and FD-GAN~\cite{ge2018fd} on all five adopted datasets, their results are obtained by running the released code with the default implementation setup. For SING~\cite{jiao2018deep}, we reproduce their results on MLR-Market-$1501$ and MLR-DukeMTMC-reID.


We note that the performance of our method can be further improved by applying pre-/post-processing methods, attention mechanisms, or re-ranking. For fair comparisons, no such techniques are used in all of our experiments.


\vspace{\paramargin}
\paragraph{Results.} In Table~\ref{table:exp-ReID}, our method performs favorably against all competing methods on all five datasets. We observe that our method consistently outperforms the best competitors~\cite{ge2018fd,wang2018cascaded} by $4\%\sim8\%$ at rank $1$.
The performance gains can be ascribed to three main factors. First, unlike most existing person re-ID methods, our model performs cross-resolution person re-ID in an end-to-end learning fashion. Second, our method learns resolution-invariant representations, allowing our model to recognize persons in images of different resolutions. Third, our model learns to recover the missing details in LR input images, thus providing additional discriminative evidence for person re-ID.

The advantage of deriving joint representation $\mathbfit{v} = [f,g]$ can be assessed by comparing with two of our variant methods, \ie Ours ($f$ only) and Ours ($g$ only). In method ``Ours ($f$ only)'', the classifier $\mathcal{C}$ only takes the resolution-invariant representation $f$ as input. In method ``Ours ($g$ only)'', the classifier $\mathcal{C}$ only takes the HR representation $g$ as input.
We observe that deriving joint representation $\mathbfit{v}$ consistently improves the performance over these two baseline methods. We note that method ``Ours ($f$ only)'' achieves a better performance than method ``Ours ($g$ only)'' on the CAVIAR dataset. We attribute the results to the higher resolution variations exhibited in the CAVIAR dataset.




\input{exp/psnr.tex}

\vspace{-3.5mm}

\input{exp/GAN}

\subsection{Evaluation of the Recovered HR Images}

To demonstrate that our CRGAN is capable of recovering the missing details in LR images of varying and even unseen resolutions, we evaluate the quality of the recovered HR images on the MLR-CUHK03 \emph{test set} using SSIM, PSNR, and LPIPS~\cite{zhang2018unreasonable} metrics. We employ the ImageNet-pretrained AlexNet~\cite{krizhevsky2012imagenet} when computing LPIPS. We compare our CRGAN with CycleGAN~\cite{zhu2017unpaired}, SING~\cite{jiao2018deep}, and CSR-GAN~\cite{wang2018cascaded}. For CycleGAN~\cite{zhu2017unpaired}, we train the model to learn a mapping between LR and HR images. We report the quantitative results of the recovered image quality and person re-ID in Table~\ref{table:image-comparison} with two different settings: (1) LR images of resolutions seen during training, \ie $r \in \{2, 3, 4\}$, and (2) LR images of unseen resolution, \ie $r = 8$.

For seen resolutions (\ie left block), we observe that our results using SSIM and PSNR metrics are slightly worse than CSR-GAN~\cite{wang2018cascaded} while compares favorably against SING~\cite{jiao2018deep} and CycleGAN~\cite{zhu2017unpaired}. However, our method performs favorably against these three methods using LPIPS metric and achieves the state-of-the-art performance when evaluating on cross-resolution person re-ID task. These results indicate that (1) SSIM and PSNR metrics are low-level pixel-wise metrics, which do not reflect high-level perceptual tasks and (2) the end-to-end learning of cross-resolution person re-ID would result in better person re-ID performance and recover more perceptually realistic HR images as reflected by LPIPS. 

For unseen resolution (\ie right block), our method performs favorably against all three competing methods on all the adopted evaluation metrics. These results suggest that our method is capable of handling unseen resolution (\ie $r = 8$) with favorable performance in terms of both image quality and person re-ID. Note that we only train our model with HR images and LR ones with $r \in \{2, 3, 4\}$.

Figure~\ref{fig:HR-img} presents six examples. For each person, there are four different resolutions (\ie $r \in \{1, 2, 4, 8\}$). Note that images with down-sampling rate $r = 1$ indicate that the images remain their original sizes and are the corresponding HR images of the LR ones. We observe that when LR images with down-sampling rate $r = 8$ are given, our model recovers the HR details with the highest visual quality among all competing methods. 
Both quantitative and qualitative results above confirm that our model can handle \emph{a range of} seen resolutions and generalize well to \emph{unseen} resolutions using just one single model, \ie CRGAN.

\input{tsne.tex}

\subsection{Ablation Study}

To analyze the importance of each developed loss function, we conduct an ablation study on the MLR-CUHK03 dataset. Table~\ref{table:exp-abla} reports the quality of the recovered HR images and the performance of cross-resolution person re-ID recorded at rank $1$.

\vspace{-2.5mm}
\paragraph{Image-level adversarial loss $\mathcal{L}_\mathrm{adv}^{\mathcal{D}_{I}}$.} When loss $\mathcal{L}_\mathrm{adv}^{\mathcal{D}_{I}}$ is turned off, our model is not encouraged to produce perceptually realistic HR images as reflected by LPIPS, resulting in a performance drop of $2.3\%$ at rank $1$.

\vspace{-2.5mm}
\paragraph{Feature-level adversarial loss $\mathcal{L}_\mathrm{adv}^{\mathcal{D}_{F}}$.} Without loss $\mathcal{L}_\mathrm{adv}^{\mathcal{D}_{F}}$, our model does not learn resolution-invariant representations and thus suffers from the resolution mismatch issue. Significant performance drops in the recovered image quality and person re-ID performance occur, indicating the importance of our method for learning resolution-invariant representations to address the resolution mismatch issue.

\vspace{-2.5mm}
\paragraph{HR reconstruction loss $\mathcal{L}_\mathrm{rec}$.} Once loss $\mathcal{L}_\mathrm{rec}$ is excluded, there is no explicit supervision to guide the CRGAN to perform image recovery, and the model implicitly suffers from information loss in compressing visual images into semantic feature maps. Severe performance drops in terms of the recovered image quality and person re-ID performance are hence caused.

\vspace{-2.5mm}
\paragraph{Classification loss $\mathcal{L}_\mathrm{cls}$.} Although our model is still able to perform image recovery without loss $\mathcal{L}_\mathrm{cls}$, our model cannot perform discriminative learning for person re-ID since data labels are not used during training. Thus, significant performance drop in person re-ID occurs.

The ablation study demonstrates that the losses $\mathcal{L}_\mathrm{adv}^{\mathcal{D}_{F}}$, $\mathcal{L}_\mathrm{rec}$, and $\mathcal{L}_\mathrm{cls}$ are crucial to our method, while the loss $\mathcal{L}_\mathrm{adv}^{\mathcal{D}_{I}}$ is helpful for improving the performance of cross-resolution person re-ID as well as the quality of the recovered images.

\subsection{Resolution-Invariant Representation $f$}

To demonstrate the effectiveness of our model in deriving the resolution-invariant representations, we first apply global average pooling to $f$ to obtain the resolution-invariant feature vector $\mathbfit{w} = \mathrm{GAP}(f) \in \bbR^d$. We then visualize $\mathbfit{w}$ on the MLR-CUHK03 \emph{test set} in Figure~\ref{fig:tsne}.

To be more precise, we select $15$ different identities, each of which is indicated by a unique color, as shown in Figure~\ref{fig:tsne-baseline} and Figure~\ref{fig:tsne-identity}. In Figure~\ref{fig:tsne-baseline}, we observe that without the feature-level adversarial loss $\mathcal{L}_\mathrm{adv}^{\mathcal{D}_F}$, our model cannot establish a well-separated feature space. When loss $\mathcal{L}_\mathrm{adv}^{\mathcal{D}_F}$ is imposed, as shown in Figure~\ref{fig:tsne-identity}, the projected feature vectors are well separated. These two figures indicate that without loss $\mathcal{L}_\mathrm{adv}^{\mathcal{D}_F}$, our model does not learn resolution-invariant representations, thus implicitly suffering from the negative impact induced by the resolution mismatch issue.

We note that the projected feature vectors in Figure~\ref{fig:tsne-identity} are well separated, suggesting that sufficient person re-ID ability can be exhibited by our model. On the other hand, for Figure~\ref{fig:tsne-resolution}, we colorize each image resolution with a unique color in each identity cluster (four different down-sampling rates $r \in \{1, 2, 4, 8\}$). We observe that the projected feature vectors of the same identity but different down-sampling rates are all well clustered. We note that images with down-sampling rate $r = 8$ are not present in the training set (\ie unseen resolution).

The above visualizations demonstrate that our model learns resolution-invariant representations and generalizes well to unseen image resolution (\eg $r = 8$) for cross-resolution person re-ID.

\input{exp/exp-ablation-hrgan.tex}

%% file: exp/exp-ReID.tex
\begin{table*}[!htbp]
  \small
  \ra{1.3}
  \begin{center}
  \caption{\textbf{Results of cross-resolution re-ID (\%).} Bold and underlined numbers indicate top two results, respectively.}
  \vspace{-2.5mm}
  \label{table:exp-ReID}
  \resizebox{\linewidth}{!} 
  {
  \begin{tabular}{l|ccc|ccc|ccc|ccc|ccc}
  \toprule
  \multirow{2}{*}{Method} & \multicolumn{3}{c|}{MLR-CUHK03} & \multicolumn{3}{c|}{MLR-VIPeR} & \multicolumn{3}{c|}{CAVIAR} & \multicolumn{3}{c|}{MLR-Market-1501} & \multicolumn{3}{c}{MLR-DukeMTMC-reID}\\
  & Rank 1 & Rank 5 & Rank 10 & Rank 1 & Rank 5 & Rank 10 & Rank 1 & Rank 5 & Rank 10 & Rank 1 & Rank 5 & Rank 10 & Rank 1 & Rank 5 & Rank 10 \\
  \midrule
  JUDEA~\cite{li2015multi} & 26.2 & 58.0 & 73.4 & 26.0 & 55.1 & 69.2 & 22.0 & 60.1 & 80.8 & - & - & - & - & - & - \\
  SLD$^2$L~\cite{jing2015super} & - & - & -& 20.3 & 44.0 & 62.0 & 18.4 & 44.8 & 61.2  & - & - & - & - & - & - \\
  SDF~\cite{wang2016scale} & 22.2 & 48.0 & 64.0 & 9.3 & 38.1 & 52.4 & 14.3 & 37.5 & 62.5  & - & - & - & - & - & - \\
  SING~\cite{jiao2018deep} & 67.7 & 90.7 & 94.7 & 33.5 & 57.0 & 66.5 & 33.5 & 72.7 & 89.0  & 74.4 & 87.8 & 91.6 & 65.2 & 80.1 & 84.8 \\
  CSR-GAN~\cite{wang2018cascaded} & 71.3 & 92.1 & 97.4 & 37.2 & 62.3 & 71.6 & 34.7 & 72.5 & 87.4  & 76.4 & 88.5 & 91.9 & 67.6 & 81.4 & 85.1 \\
  \midrule
  CamStyle~\cite{zhong2017camera} & 69.1 & 89.6 & 93.9 & 34.4 & 56.8 & 66.6 & 32.1 & 72.3 & 85.9  & 74.5 & 88.6 & 93.0 & 64.0 & 78.1 & 84.4 \\
  FD-GAN~\cite{ge2018fd} &73.4 & 93.8 & 97.9 & 39.1 & 62.1 & 72.5 & 33.5 & 71.4 & 86.5 & 79.6 & \underline{91.6} & 93.5 & 67.5 & 82.0 & 85.3 \\
  \midrule
  Ours ($f$ only) & 77.6 & \underline{96.2} & 98.5& 41.2 & 66.3 & 75.6 & \underline{41.5} & \underline{75.3} & 85.6  & 80.1 & 90.6 & 93.2 & 73.4 & 84.4 & 86.8 \\
  Ours ($g$ only) &\underline{79.7} & \textbf{97.4} & \underline{98.7}& \underline{41.7} & \underline{66.4} & \underline{76.1} & 38.9 & 73.1 & \underline{90.6}  & \underline{82.2} & 91.3 & \underline{94.5} & \underline{74.1} & \underline{85.1} & \underline{88.2} \\
  %
  %
  Ours &\textbf{82.1} & \textbf{97.4} & \textbf{98.8} & \textbf{43.1} & \textbf{68.2} & \textbf{77.5}& \textbf{42.8} & \textbf{76.2} & \textbf{91.5}  & \textbf{83.7} & \textbf{92.7} & \textbf{95.8} & \textbf{75.6} & \textbf{86.7} & \textbf{89.6} \\
  
  \bottomrule
  \end{tabular}
  }
  \end{center}
  \vspace{-5.5mm}
\end{table*}

%% file: exp/psnr.tex
\begin{table*}[t]
  \scriptsize
  \caption{\textbf{Quantitative results of cross-resolution person re-ID on the MLR-CUHK03 test set.} \emph{Left block}: resolutions are seen during training. \emph{Right block}: resolution is not seen during training.}
  \vspace{-2.0mm}
  \centering
  \label{table:image-comparison}
  \resizebox{\linewidth}{!}
  {
  \begin{tabular}{l|ccc|c|ccc|c}
  \toprule
  \multirow{2}{*}{Method} & \multicolumn{4}{c|}{Down-sampling rate $r \in \{2, 3, 4\}$ (seen)} & \multicolumn{4}{c}{Down-sampling rate $r = 8$ (unseen)}\\\cmidrule{2-9} 
  & SSIM $\uparrow$ & PSNR $\uparrow$ & LPIPS~\cite{zhang2018unreasonable} $\downarrow$ & Rank 1 (\%) $\uparrow$ & SSIM $\uparrow$ & PSNR $\uparrow$ & LPIPS~\cite{zhang2018unreasonable} $\downarrow$ & Rank 1 (\%) $\uparrow$ \\
  \midrule
  CycleGAN~\cite{zhu2017unpaired} & 0.55 & 14.1 & 0.31 & 62.1 & 0.42 & 12.7 & 0.37 & 40.5\\
  SING~\cite{jiao2018deep} & 0.65 & 18.1 & 0.18 & 67.7 & 0.52 & 14.5 & 0.34 & 54.2\\
  CSR-GAN~\cite{wang2018cascaded} & \textbf{0.76} & \textbf{21.5} & 0.13 & 71.3 & 0.67 & 17.2 & 0.25 & 62.1\\
  %
  %
  \midrule
  Ours & 0.73 & 20.2 & \textbf{0.07} & \textbf{82.1} & \textbf{0.71} & \textbf{19.8} & \textbf{0.11} & \textbf{78.6}\\
  \bottomrule
  \end{tabular}
  }
\end{table*}

%% file: exp/GAN.tex
\begin{figure*}[t]
  \centering
  \includegraphics[width=\linewidth]{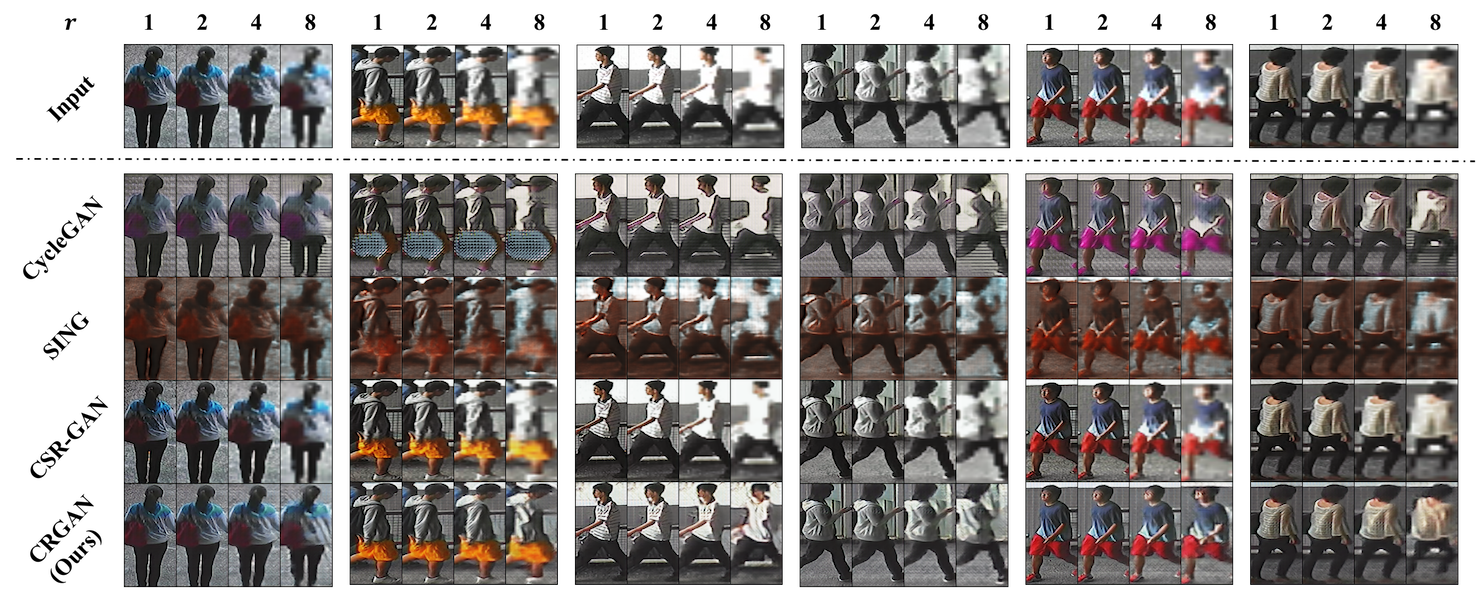}
  \vspace{-5.5mm}
  \caption{\textbf{Visual results of the recovered HR images on the MLR-CUHK03 test set.} We present the visual comparison among CycleGAN~\cite{zhu2017unpaired}, SING~\cite{jiao2018deep}, CSR-GAN~\cite{wang2018cascaded}, and the proposed CRGAN.}
  \vspace{-3.5mm}
  \label{fig:HR-img}
\end{figure*}

%% file: tsne.tex
\setlength{\threeimg}{0.323\textwidth}
\begin{figure*}[t]
  \centering
  \begin{subfigure}[b]{\threeimg}
    \centering
    \includegraphics[width=\linewidth]{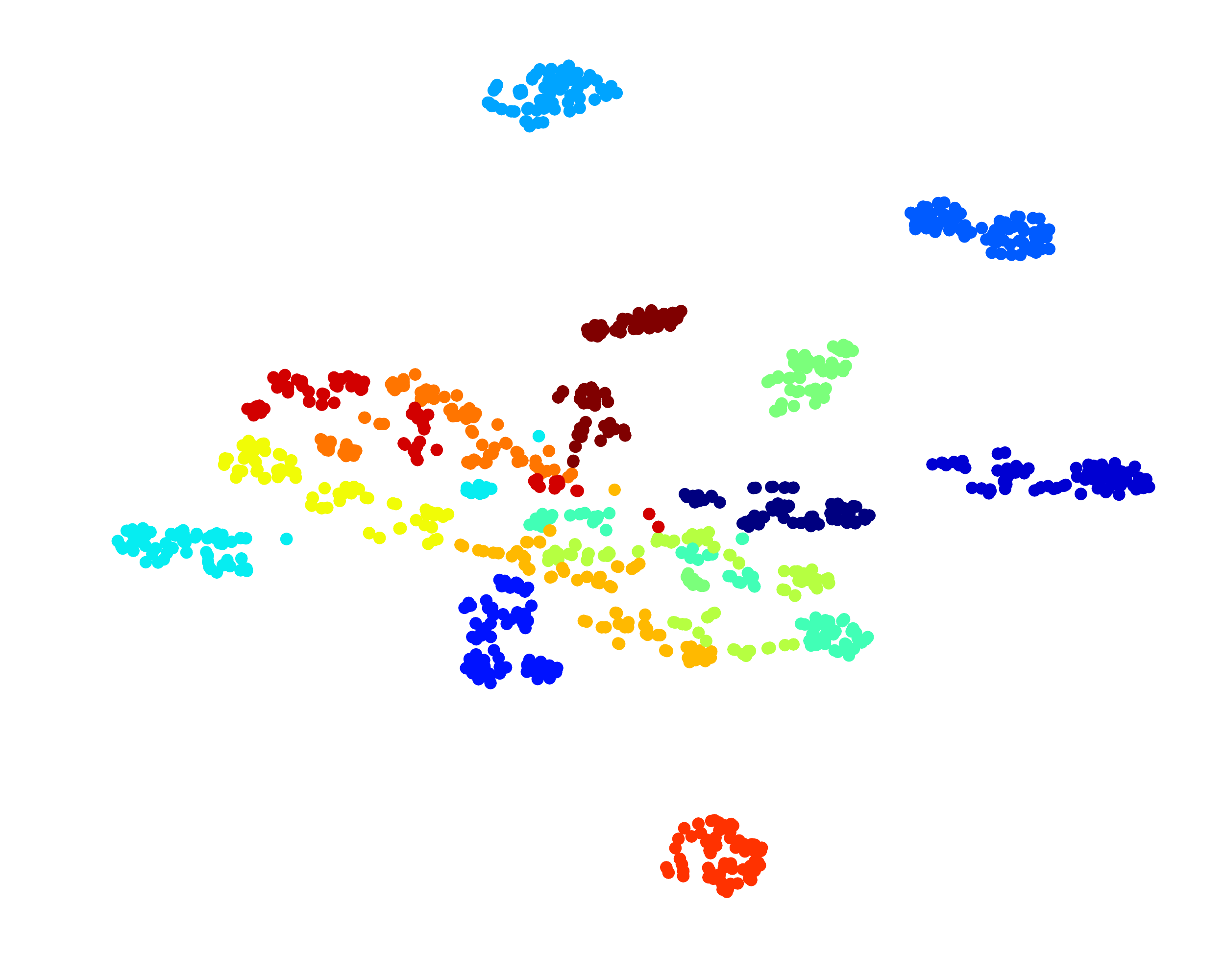}
    \vspace{-6.5mm}
    \caption{Ours w/o $\mathcal{L}_\mathrm{adv}^{\mathcal{D}_{F}}$: colorized w.r.t \textbf{identity}.}
    \label{fig:tsne-baseline}
  \end{subfigure}
  \hspace{1.0mm}
  \begin{subfigure}[b]{\threeimg}
    \centering
    \includegraphics[width=\linewidth]{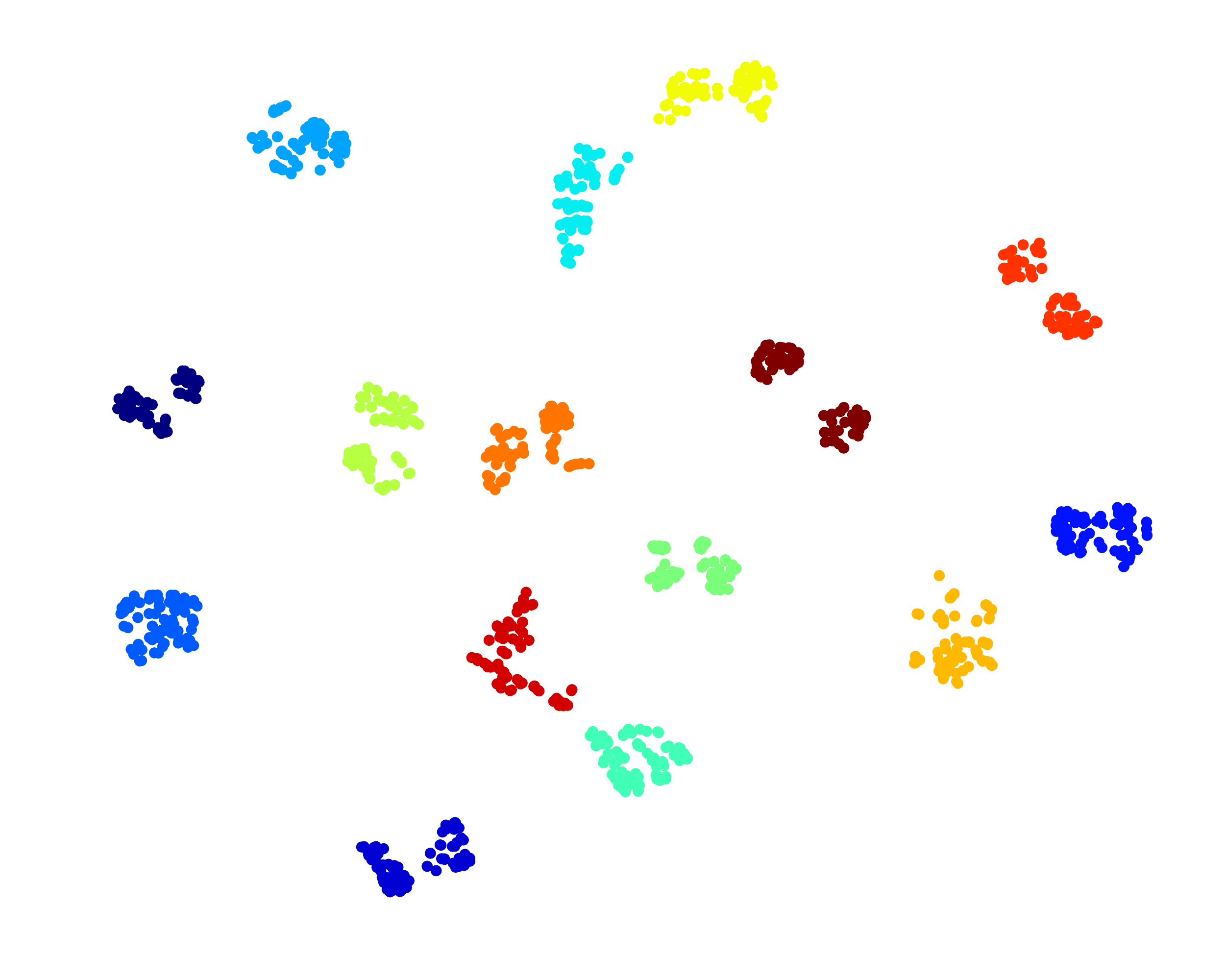}
    \vspace{-6.5mm}
    \caption{Ours: colorized w.r.t \textbf{identity}.}
    \label{fig:tsne-identity}
  \end{subfigure}
  \hspace{1.0mm}
  \begin{subfigure}[b]{\threeimg}
    \centering
    \includegraphics[width=\linewidth]{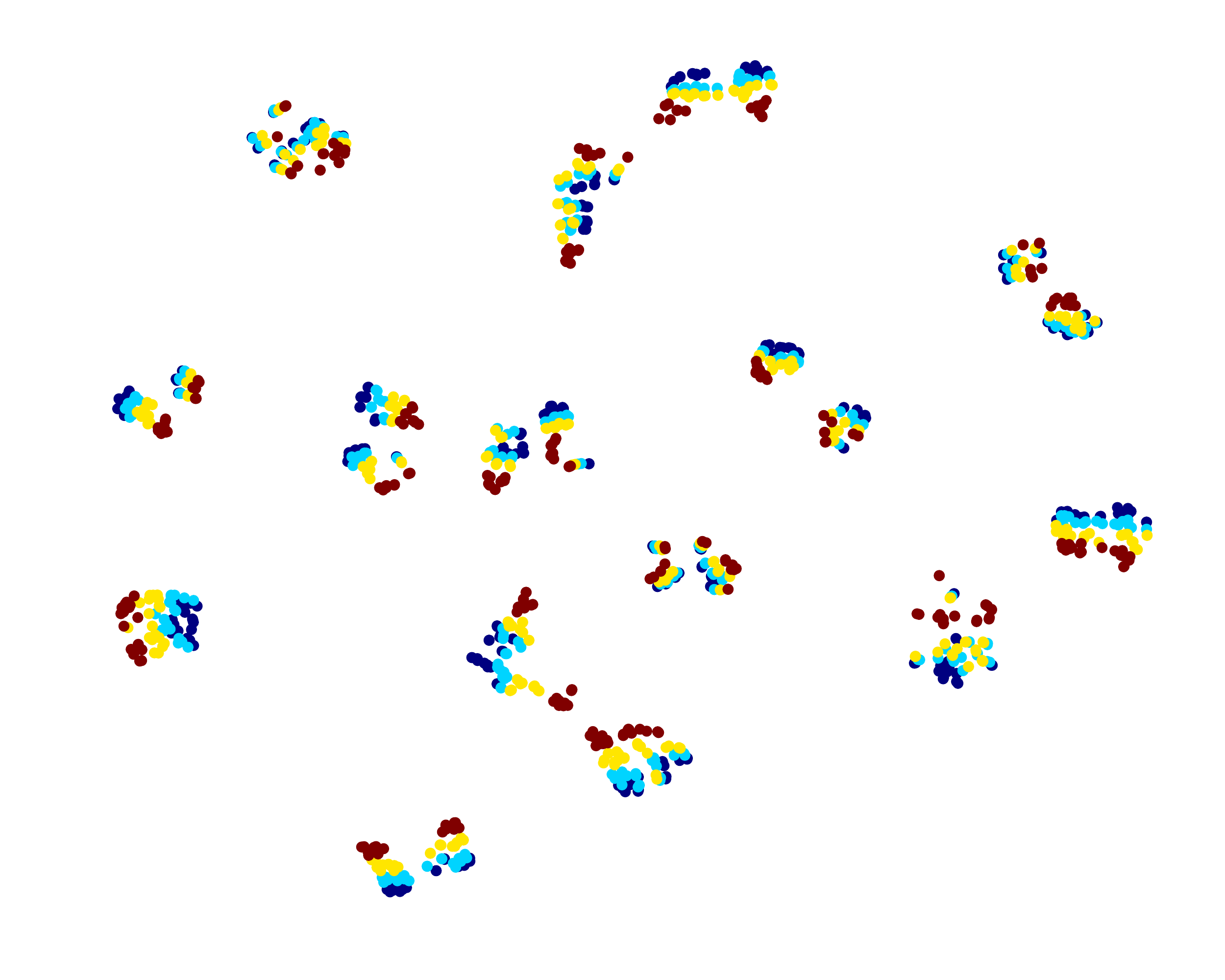}
    \vspace{-6.5mm}
    \caption{Ours: colorized w.r.t \textbf{resolution}.}
    \label{fig:tsne-resolution}
  \end{subfigure}
  \vspace{-5.0mm}
  \caption{\textbf{2D visualization of the resolution-invariant feature vector $\mathbfit{w}$ on the MLR-CUHK03 test set via t-SNE.} Data of different identities (each in a unique color) derived by our model \emph{without} and \emph{with} observing the feature-level adversarial loss $\mathcal{L}_\mathrm{adv}^{\mathcal{D}_{F}}$ are shown in (a) and (b), respectively. The same data but with resolution-specific colorization, \ie one color for each down-sampling rate $r \in \{1, 2, 4, 8\}$, are depicted in (c). Note that images with $r = 8$ are not seen during training.}
  \vspace{-3.0mm}
  \label{fig:tsne}
\end{figure*}

%% file: exp/exp-ablation-hrgan.tex
\begin{table}[t]
  \small
  \ra{1.3}
  \begin{center}
  \caption{\textbf{Ablation study on the MLR-CUHK03 dataset.} Bold and underlined numbers indicate top two results, respectively.}
  \vspace{-2.5mm}
  \label{table:exp-abla}
  \resizebox{\linewidth}{!} 
  {
  \begin{tabular}{l|ccc|c}
  \toprule
  %
  %
  Method & SSIM $\uparrow$ & PSNR $\uparrow$  & LPIPS~\cite{zhang2018unreasonable} $\downarrow$ & Rank 1 (\%) $\uparrow$ \\
  \midrule
  Ours & \textbf{0.73} & \underline{20.2} & \textbf{0.07} & \textbf{82.1} \\
  Ours w/o $\mathcal{L}_\mathrm{adv}^{\mathcal{D}_{I}}$& 0.67 & 18.5 & 0.17 & \underline{79.8} \\
  Ours w/o $\mathcal{L}_\mathrm{adv}^{\mathcal{D}_{F}}$& 0.54 & 14.2 & 0.34 & 67.6 \\
  Ours w/o $\mathcal{L}_\mathrm{rec}$& 0.45 & 12.9 & 0.40 & 66.7 \\
  Ours w/o $\mathcal{L}_\mathrm{cls}$ & \underline{0.72} & \textbf{21.4} & \underline{0.11} & 1.7 \\
  \bottomrule
  \end{tabular}
  }
  \end{center}
  \vspace{-4.0mm}
\end{table}

%% file: conclusion.tex
\section{Conclusions}

We have presented an \emph{end-to-end trainable} generative adversarial network, CAD-Net, for addressing the resolution mismatch issue in person re-ID. The core technical novelty lies in the unique design of the proposed CRGAN which learns the \emph{resolution-invariant} representations while being able to recover \emph{re-ID oriented} HR details. Our cross-modal re-ID network jointly considers the information from two feature modalities, leading to better person re-ID capability. Extensive experimental results demonstrate that our approach performs favorably against existing cross-resolution and standard person re-ID methods on five challenging benchmarks, and produces perceptually higher quality HR images using only a \emph{single} model. Visualization of the resolution-invariant representations further verifies our ability in handling query images with \emph{varying} or even \emph{unseen} resolutions. Thus, the use of our model for practical person re-ID applications can be strongly supported.

%% file: main.bbl
\begin{thebibliography}{10}\itemsep=-1pt

\bibitem{andriluka2008people}
Mykhaylo Andriluka, Stefan Roth, and Bernt Schiele.
\newblock People-tracking-by-detection and people-detection-by-tracking.
\newblock In {\em CVPR}, 2008.

\bibitem{chang2018multi}
Xiaobin Chang, Timothy~M Hospedales, and Tao Xiang.
\newblock Multi-level factorisation net for person re-identification.
\newblock In {\em CVPR}, 2018.

\bibitem{chen2018group}
Dapeng Chen, Dan Xu, Hongsheng Li, Nicu Sebe, and Xiaogang Wang.
\newblock Group consistent similarity learning via deep crf for person
  re-identification.
\newblock In {\em CVPR}, 2018.

\bibitem{chen2019saliency}
Yun-Chun Chen and Winston~H Hsu.
\newblock Saliency aware: Weakly supervised object localization.
\newblock In {\em ICASSP}, 2019.

\bibitem{chen2018deep}
Yun-Chun Chen, Po-Hsiang Huang, Li-Yu Yu, Jia-Bin Huang, Ming-Hsuan Yang, and
  Yen-Yu Lin.
\newblock Deep semantic matching with foreground detection and
  cycle-consistency.
\newblock In {\em ACCV}, 2018.

\bibitem{RAIN}
Yun-Chun Chen, Yu-Jhe Li, Xiaofei Du, and Yu-Chiang~Frank Wang.
\newblock Learning resolution-invariant deep representations for person
  re-identification.
\newblock In {\em AAAI}, 2019.

\bibitem{chen2017deep}
Yun-Chun Chen, Yu-Jhe Li, Aragorn Tseng, and Tsungnan Lin.
\newblock Deep learning for malicious flow detection.
\newblock In {\em PIMRC}, 2017.

\bibitem{chen2019crdoco}
Yun-Chun Chen, Yen-Yu Lin, Ming-Hsuan Yang, and Jia-Bin Huang.
\newblock Crdoco: Pixel-level domain transfer with cross-domain consistency.
\newblock In {\em CVPR}, 2019.

\bibitem{chen2019show}
Yun-Chun Chen, Yen-Yu Lin, Ming-Hsuan Yang, and Jia-Bin Huang.
\newblock Show, match and segment: Joint learning of semantic matching and
  object co-segmentation.
\newblock {\em arXiv}, 2019.

\bibitem{cheng2016person}
De Cheng, Yihong Gong, Sanping Zhou, Jinjun Wang, and Nanning Zheng.
\newblock Person re-identification by multi-channel parts-based cnn with
  improved triplet loss function.
\newblock In {\em CVPR}, 2016.

\bibitem{Cheng:BMVC11}
D.~S. Cheng, M. Cristani, M. Stoppa, L. Bazzani, and V. Murino.
\newblock Custom pictorial structures for re-identification.
\newblock In {\em BMVC}, 2011.

\bibitem{dahl2017pixel}
Ryan Dahl, Mohammad Norouzi, and Jonathon Shlens.
\newblock Pixel recursive super resolution.
\newblock In {\em ICCV}, 2017.

\bibitem{image-image18}
Weijian Deng, Liang Zheng, Qixiang Ye, Guoliang Kang, Yi Yang, and Jianbin
  Jiao.
\newblock Image-image domain adaptation with preserved self-similarity and
  domain-dissimilarity for person reidentification.
\newblock In {\em CVPR}, 2018.

\bibitem{dong2016image}
Chao Dong, Chen~Change Loy, Kaiming He, and Xiaoou Tang.
\newblock Image super-resolution using deep convolutional networks.
\newblock {\em TPAMI}, 2016.

\bibitem{ge2018fd}
Yixiao Ge, Zhuowan Li, Haiyu Zhao, Guojun Yin, Shuai Yi, Xiaogang Wang, and
  Hongsheng Li.
\newblock Fd-gan: Pose-guided feature distilling gan for robust person
  re-identification.
\newblock In {\em NeurIPS}, 2018.

\bibitem{goodfellow2014generative}
Ian Goodfellow, Jean Pouget-Abadie, Mehdi Mirza, Bing Xu, David Warde-Farley,
  Sherjil Ozair, Aaron Courville, and Yoshua Bengio.
\newblock Generative adversarial nets.
\newblock In {\em NeurIPS}, 2014.

\bibitem{gray2008viewpoint}
Douglas Gray and Hai Tao.
\newblock Viewpoint invariant pedestrian recognition with an ensemble of
  localized features.
\newblock In {\em ECCV}, 2008.

\bibitem{he2016deep}
Kaiming He, Xiangyu Zhang, Shaoqing Ren, and Jian Sun.
\newblock Deep residual learning for image recognition.
\newblock In {\em CVPR}, 2016.

\bibitem{hermans2017defense}
Alexander Hermans, Lucas Beyer, and Bastian Leibe.
\newblock In defense of the triplet loss for person re-identification.
\newblock {\em arXiv}, 2017.

\bibitem{hoffman2017cycada}
Judy Hoffman, Eric Tzeng, Taesung Park, Jun-Yan Zhu, Phillip Isola, Kate
  Saenko, Alexei~A Efros, and Trevor Darrell.
\newblock Cycada: Cycle-consistent adversarial domain adaptation.
\newblock In {\em ICML}, 2018.

\bibitem{huang2018munit}
Xun Huang, Ming-Yu Liu, Serge Belongie, and Jan Kautz.
\newblock Multimodal unsupervised image-to-image translation.
\newblock In {\em ECCV}, 2018.

\bibitem{jiao2018deep}
Jiening Jiao, Wei-Shi Zheng, Ancong Wu, Xiatian Zhu, and Shaogang Gong.
\newblock Deep low-resolution person re-identification.
\newblock In {\em AAAI}, 2018.

\bibitem{jing2015super}
Xiao-Yuan Jing, Xiaoke Zhu, Fei Wu, Xinge You, Qinglong Liu, Dong Yue, Ruimin
  Hu, and Baowen Xu.
\newblock Super-resolution person re-identification with semi-coupled low-rank
  discriminant dictionary learning.
\newblock In {\em CVPR}, 2015.

\bibitem{kalayeh2018human}
Mahdi~M Kalayeh, Emrah Basaran, Muhittin G{\"o}kmen, Mustafa~E Kamasak, and
  Mubarak Shah.
\newblock Human semantic parsing for person re-identification.
\newblock In {\em CVPR}, 2018.

\bibitem{khan2016person}
Furqan~M Khan and Fran{\c{c}}ois Br{\'e}mond.
\newblock Person re-identification for real-world surveillance systems.
\newblock {\em arXiv}, 2016.

\bibitem{kim2016accurate}
Jiwon Kim, Jung Kwon~Lee, and Kyoung Mu~Lee.
\newblock Accurate image super-resolution using very deep convolutional
  networks.
\newblock In {\em CVPR}, 2016.

\bibitem{krizhevsky2012imagenet}
Alex Krizhevsky, Ilya Sutskever, and Geoffrey~E Hinton.
\newblock Imagenet classification with deep convolutional neural networks.
\newblock In {\em NeurIPS}, 2012.

\bibitem{ledig2017photo}
Christian Ledig, Lucas Theis, Ferenc Husz{\'a}r, Jose Caballero, Andrew
  Cunningham, Alejandro Acosta, Andrew~P Aitken, Alykhan Tejani, Johannes Totz,
  Zehan Wang, et~al.
\newblock Photo-realistic single image super-resolution using a generative
  adversarial network.
\newblock In {\em CVPR}, 2017.

\bibitem{li2014deepreid}
Wei Li, Rui Zhao, Tong Xiao, and Xiaogang Wang.
\newblock Deepreid: Deep filter pairing neural network for person
  re-identification.
\newblock In {\em CVPR}, 2014.

\bibitem{li2018harmonious}
Wei Li, Xiatian Zhu, and Shaogang Gong.
\newblock Harmonious attention network for person re-identification.
\newblock In {\em CVPR}, 2018.

\bibitem{li2015multi}
Xiang Li, Wei-Shi Zheng, Xiaojuan Wang, Tao Xiang, and Shaogang Gong.
\newblock Multi-scale learning for low-resolution person re-identification.
\newblock In {\em ICCV}, 2015.

\bibitem{liao2015person}
Shengcai Liao, Yang Hu, Xiangyu Zhu, and Stan~Z Li.
\newblock Person re-identification by local maximal occurrence representation
  and metric learning.
\newblock In {\em CVPR}, 2015.

\bibitem{lin2018learning}
Jhih-Yuan Lin, Min-Sheng Wu, Yu-Cheng Chang, Yun-Chun Chen, Chao-Te Chou,
  Chun-Ting Wu, and Winston~H Hsu.
\newblock Learning volumetric segmentation for lung tumor.
\newblock {\em IEEE ICIP VIP Cup Tech. Report}, 2018.

\bibitem{lin2017improving}
Yutian Lin, Liang Zheng, Zhedong Zheng, Yu Wu, and Yi Yang.
\newblock Improving person re-identification by attribute and identity
  learning.
\newblock {\em arXiv}, 2017.

\bibitem{liu2018pose}
Jinxian Liu, Bingbing Ni, Yichao Yan, Peng Zhou, Shuo Cheng, and Jianguo Hu.
\newblock Pose transferrable person re-identification.
\newblock In {\em CVPR}, 2018.

\bibitem{miyato2018cgans}
Takeru Miyato and Masanori Koyama.
\newblock cgans with projection discriminator.
\newblock In {\em ICLR}, 2018.

\bibitem{shen2018deep}
Yantao Shen, Hongsheng Li, Tong Xiao, Shuai Yi, Dapeng Chen, and Xiaogang Wang.
\newblock Deep group-shuffling random walk for person re-identification.
\newblock In {\em CVPR}, 2018.

\bibitem{shen2018person}
Yantao Shen, Hongsheng Li, Shuai Yi, Dapeng Chen, and Xiaogang Wang.
\newblock Person re-identification with deep similarity-guided graph neural
  network.
\newblock In {\em ECCV}, 2018.

\bibitem{si2018dual}
Jianlou Si, Honggang Zhang, Chun-Guang Li, Jason Kuen, Xiangfei Kong, Alex~C
  Kot, and Gang Wang.
\newblock Dual attention matching network for context-aware feature sequence
  based person re-identification.
\newblock In {\em CVPR}, 2018.

\bibitem{song2018mask}
Chunfeng Song, Yan Huang, Wanli Ouyang, and Liang Wang.
\newblock Mask-guided contrastive attention model for person re-identification.
\newblock In {\em CVPR}, 2018.

\bibitem{tsai2018learning}
Yi-Hsuan Tsai, Wei-Chih Hung, Samuel Schulter, Kihyuk Sohn, Ming-Hsuan Yang,
  and Manmohan Chandraker.
\newblock Learning to adapt structured output space for semantic segmentation.
\newblock In {\em CVPR}, 2018.

\bibitem{vezzani2013people}
Roberto Vezzani, Davide Baltieri, and Rita Cucchiara.
\newblock People reidentification in surveillance and forensics: A survey.
\newblock {\em ACM Computing Surveys (CSUR)}, 2013.

\bibitem{wang2016scale}
Zheng Wang, Ruimin Hu, Yi Yu, Junjun Jiang, Chao Liang, and Jinqiao Wang.
\newblock Scale-adaptive low-resolution person re-identification via learning a
  discriminating surface.
\newblock In {\em IJCAI}, 2016.

\bibitem{wang2018cascaded}
Zheng Wang, Mang Ye, Fan Yang, Xiang Bai, and Shin'ichi Satoh.
\newblock Cascaded sr-gan for scale-adaptive low resolution person
  re-identification.
\newblock In {\em IJCAI}, 2018.

\bibitem{wei2018person}
Longhui Wei, Shiliang Zhang, Wen Gao, and Qi Tian.
\newblock Person transfer gan to bridge domain gap for person
  re-identification.
\newblock In {\em CVPR}, 2018.

\bibitem{yu2017hallucinating}
Xin Yu and Fatih Porikli.
\newblock Hallucinating very low-resolution unaligned and noisy face images by
  transformative discriminative autoencoders.
\newblock In {\em CVPR}, 2017.

\bibitem{zhang2018unreasonable}
Richard Zhang, Alexei~A Efros, Eli Shechtman, and Oliver Wang.
\newblock The unreasonable effectiveness of deep features as a perceptual
  metric.
\newblock In {\em CVPR}, 2018.

\bibitem{zheng2015scalable}
Liang Zheng, Liyue Shen, Lu Tian, Shengjin Wang, Jingdong Wang, and Qi Tian.
\newblock Scalable person re-identification: A benchmark.
\newblock In {\em ICCV}, 2015.

\bibitem{zheng2016person}
Liang Zheng, Yi Yang, and Alexander~G Hauptmann.
\newblock Person re-identification: Past, present and future.
\newblock {\em arXiv}, 2016.

\bibitem{zheng2017unlabeled}
Zhedong Zheng, Liang Zheng, and Yi Yang.
\newblock Unlabeled samples generated by gan improve the person
  re-identification baseline in vitro.
\newblock In {\em ICCV}, 2017.

\bibitem{zhong2017camera}
Zhun Zhong, Liang Zheng, Zhedong Zheng, Shaozi Li, and Yi Yang.
\newblock Camera style adaptation for person re-identification.
\newblock In {\em CVPR}, 2018.

\bibitem{zhu2017unpaired}
Jun-Yan Zhu, Taesung Park, Phillip Isola, and Alexei~A Efros.
\newblock Unpaired image-to-image translation using cycle-consistent
  adversarial networks.
\newblock In {\em ICCV}, 2017.

\bibitem{zhu2016deep}
Shizhan Zhu, Sifei Liu, Chen~Change Loy, and Xiaoou Tang.
\newblock Deep cascaded bi-network for face hallucination.
\newblock In {\em ECCV}, 2016.

\end{thebibliography}
